\pdfoutput=1

\documentclass[11pt]{article}

\usepackage[]{acl}

\usepackage{times}
\usepackage{latexsym}
\usepackage{graphicx}
\usepackage{amsmath}
\usepackage{multirow}
\usepackage{booktabs} 
\usepackage{tablefootnote}
\usepackage{colortbl} 
\usepackage{paralist} 
\usepackage{soul} 
\usepackage{xspace}
\definecolor{lightgrey}{HTML}{dcdbdb}
\sethlcolor{lightgrey}
\newcommand{\cc}[0]{\cellcolor{lightgrey}}

\newcommand{\proposed}{\textsc{ZeroGen}\xspace}
\newcommand{\smallmodel}{\textsc{TAM}\xspace}
\newcommand{\smallmodels}{\textsc{TAM}s\xspace}
\newenvironment{itemizesquish}{\begin{list}{\labelitemi}{\setlength{\itemsep}{-0.3em}\setlength{\labelwidth}{0.5em}\setlength{\leftmargin}{\labelwidth}\addtolength{\leftmargin}{\labelsep}}}{\end{list}}
\usepackage[T1]{fontenc}

\usepackage[utf8]{inputenc}

\usepackage{microtype}
\usepackage{makecell}
\title{\textsc{ZeroGen}: Efficient Zero-shot Learning via Dataset Generation}

\author{
Jiacheng Ye$^{\spadesuit\diamondsuit}$\thanks{\, Work done while interning at Shanghai AI Lab.}$\;\,^\dagger$, 
Jiahui Gao$^{\spadesuit}$\thanks{\, Equal Contribution.}, 
Qintong Li$^{\spadesuit}$,
Hang Xu$^{\clubsuit}$,
Jiangtao Feng$^{\diamondsuit}$,\\
\textbf{Zhiyong Wu}$^{\diamondsuit}$,
\textbf{Tao Yu}$^{\spadesuit\heartsuit}$,
\textbf{Lingpeng Kong}$^{\spadesuit\diamondsuit}$
\\
$^\diamondsuit$Shanghai AI Laboratory \quad
$^\clubsuit$Huawei Noah’s Ark Lab \quad
$^\heartsuit$University of Washington \\
$^\spadesuit$The University of Hong Kong \\
\texttt{\{carsonye,sumiler,qtli\}@connect.hku.hk, xbjxh@live.com,} \\ 
\texttt{\{fengjiangtao,wuzhiyong\}@pjlab.org.cn, \{tyu,lpk\}@cs.hku.hk }
}

\begin{document}
\maketitle

\begin{abstract}
There is a growing interest in dataset generation recently due to the superior generative capacity of large pre-trained language models (PLMs). In this paper, we study a flexible and efficient zero-short learning method, \textsc{ZeroGen}.
Given a zero-shot task, we first generate a dataset from scratch using PLMs in an unsupervised manner. Then, we train a tiny task model (e.g., LSTM) under the supervision of the synthesized dataset. This approach allows highly efficient inference as the final task model only has orders of magnitude fewer parameters comparing to PLMs (e.g., GPT2-XL).
Apart from being annotation-free and efficient, we argue that \textsc{ZeroGen} can also provide useful insights from the perspective of data-free model-agnostic knowledge distillation, and unreferenced text generation evaluation. 
Experiments and analysis on different NLP tasks, namely, text  classification,  question  answering, and natural language inference, show the effectiveness of \textsc{ZeroGen}.


\end{abstract}

\section{Introduction}

While generating training data with language model is not new to natural language processing \cite{DBLP:conf/aaai/Anaby-TavorCGKK20,DBLP:conf/emnlp/PuriSSPC20,DBLP:journals/corr/abs-2003-02245}, it has garnered enormous interests recently due to the superior generative capacity of large-scale pre-trained language models (PLMs). Training examples created in such a manner have been found effective in various scenarios via the data augmentation procedure \citep[\emph{inter alia}]{DBLP:journals/corr/abs-2102-01335,DBLP:conf/emnlp/SchickS21a,DBLP:journals/corr/abs-2109-09193,meng2022generating}. 

In this paper, we study
an extreme scenario of such an approach, \proposed.
Given a downstream task, we first generate its training data from scratch using a powerful PLM, whose generation is steered by carefully designed task-specific prompts. Then, we train a \underline{t}iny t\underline{a}sk \underline{m}odel (\smallmodel), which has orders of magnitude fewer parameters than PLMs, under the supervision of the synthesized training data. Machine generated text is the only medium that connects the PLMs to the final task models, and no human annotations are required in the entire process. The TAM can be of any choice (e.g., loglinear or neural), allowing efficient inference\footnote{Amazon estimates that
90\% of production ML infrastructure 
costs are for inference
, rather than training \cite{DBLP:journals/corr/abs-1901-10008}.} and deployment. Besides, TAM can be flexibly designed with any task-specific strategies (e.g., inductive bias or loss), which could provide superior performance.
 
Apart from being annotation-free and efficient, we are also interested in \proposed for the following reasons. First, \proposed can be seen as a variant of knowledge distillation (KD; \citet{DBLP:journals/corr/HintonVD15}) that provides some exciting new features. Unlike conventional KD, \proposed\ does not require any human annotations during distillation. Furthermore, \proposed\ makes no presumption on the architecture choice of student models, thus we can incorporate any task-specific inductive bias into the design of student models conveniently. Second, \proposed can serve as an  unreferenced evaluation method for text generation \citep{DBLP:conf/emnlp/GuanH20,pillutla2021mauve}: the downstream tasks' performance is  dominated by the quality of the synthesized text, thus can serve as an \textit{indirect} measure of the generation models and algorithms.  
Third, \proposed sheds new lights on \emph{prompt engineering} \citep{DBLP:conf/emnlp/PetroniRRLBWM19,DBLP:conf/nips/BrownMRSKDNSSAA20} (i.e., the design of the prompts in PLMs).
As manual prompts reflect our essential knowledge of specific tasks, an intriguing question here is to what extend we can incorporate human knowledge or instructions in these prompts.

We evaluate \proposed in three NLP tasks which are text classification, question answering, and natural language inference, across six datasets. Our key research findings are summarized as follows:
\begin{itemizesquish}
    \item The zero-shot performance of \smallmodel in \proposed framework significantly surpasses its PLM counterparts (which often serves as the teacher models under the knowledge distillation context), with only $\sim$0.4\% number of parameters (\S\ref{sec:main});
    \item  In some low-resourced settings, \smallmodel trained with synthesized data even outperforms the same model trained with human annotations in a fully supervised manner (\S\ref{sec:scale});
    \item The quality of the generated text by known models and algorithms are well reflected in downstream tasks' performance, and decoding strategies that encourage more diversity also result in greater noise  (\S\ref{sec:decode});
    \item Prompt engineering is challenging -- the performance of more instructive or natural language style prompts varies in different tasks (\S\ref{sec:prompt}).
\end{itemizesquish}

In conclusion, we argue that \proposed is a viable and promising approach towards flexible and efficient zero-shot learning in NLP. It also has a great potential as a data-free model-agnostic knowledge distillation and unreferenced text evaluation method. Our code can be found at 
\url{https://github.com/HKUNLP/ZeroGen}.

\begin{figure*}[t]
\centering
\includegraphics[width=5.9in]{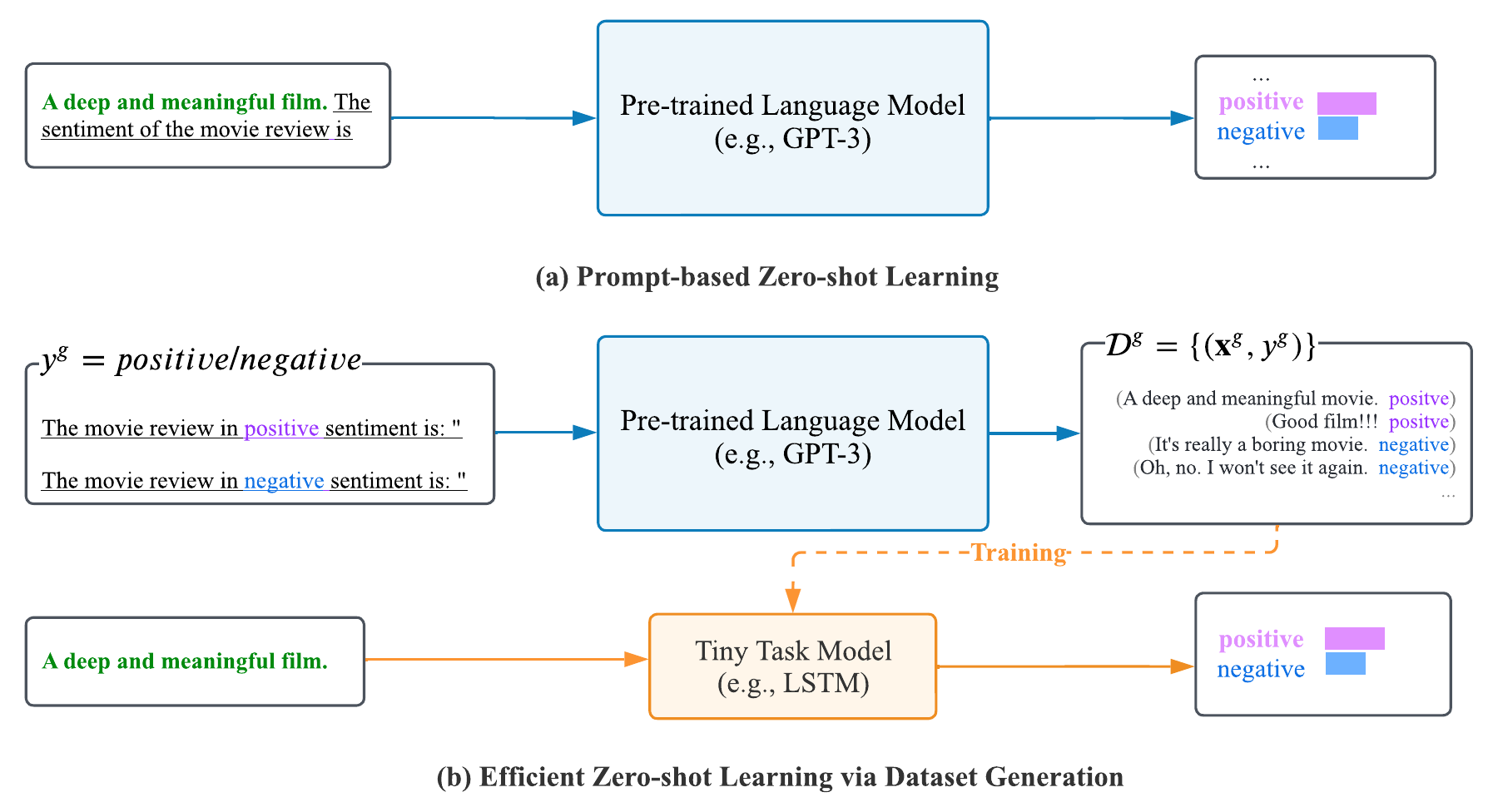}
\caption{\textbf{(a)} Prompt-based zero-shot learning (\textsc{Prompting}) framework. The text in \textcolor[HTML]{008a0e}{\textbf{green}} is the sentence to be classified. After concatenating the sentence with each prompt, a huge PLM (e.g., GPT-3) is used to calculate LM likelihood score for each class. 
\textbf{(b)} Our \textsc{ZeroGen} framework. We first generate a training set with PLM in a purely unsupervised manner. After simple filtering operations, we then train a tiny task model (e.g., LSTM) for flexible and efficient inference. The dash line indicates that the training procedure is only performed once before inference.
}
\label{fig:model}
\end{figure*}


\section{Preliminary: Prompt-based Zero-Shot Learning}
We start with preliminary knowledge about prompt-based zero-shot learning framework (named \textsc{Prompting}).

Giving a pre-trained language model (PLM) $\mathcal{P}$ and a text classification (TC) task $\mathcal{D} = (\mathcal{X}, \mathcal{Y})$, \textsc{Prompting} first instantiates a prompt $\mathcal{T(\cdot)}$ with each input $\mathbf{x}_i \in \mathcal{X}$ and outputs a natural language sequence to be completed by $\mathcal{P}$. For instance, we show an example on sentiment analysis task in Figure \ref{fig:model}(a), where $\mathbf{x}_i$ is "\textit{A deep and meaningful film.}" and $\mathcal{T}(\mathbf{x}_i)$ is "\textit{A deep and meaningful film. The sentiment of the movie review is }".
Furthermore, \textsc{Prompting} defines a verbalizer $\mathcal{M(\cdot)}$ that maps each label/class $y_i$ to a word/words in $\mathcal{P}$'s vocabulary. For instance, "\textit{positive}" and "\textit{negative}" represents the two classes. In this way, \textsc{Prompting} models the probability of class $y_i \in \mathcal{Y}$ for $\mathbf{x}_i$ as:
\begin{equation}
p(y_i|\mathbf{x}_i) = \mathcal{P}\left(\mathcal{M}(y_i) | \mathcal{T}(\mathbf{x}_i)\right).    
\end{equation}
During the whole process, the pre-trained weights of $\mathcal{P}$ are frozen and no training is required.

The vast linguistic \cite{DBLP:conf/acl/JawaharSS19,DBLP:journals/corr/abs-1901-05287,DBLP:conf/iclr/TenneyXCWPMKDBD19} and factual \cite{DBLP:conf/emnlp/PetroniRRLBWM19,DBLP:journals/tacl/JiangXAN20} knowledge encoded in PLMs' parameters is the key towards \textsc{Prompting}'s success. However, \textsc{Prompting} fails to fully exert the capacity of PLMs and heavily relies on gigantic PLMs during inference. This motivates us to explore a more flexible and efficient way of conducting zero-shot learning with PLMs. 

\section{\proposed}
In this work, we take the dataset generation method to the extreme and study \proposed, a flexible and efficient \underline{zero}-shot learning framework via dataset \underline{gen}eration.
\proposed\ framework comprises three sequential stages as shown in Figure \ref{fig:model}(b):
\begin{enumerate}\itemsep 2pt
    \item 
    The goal of the first stage is to make use of the generative power of PLMs to synthesize a dataset to solve the downstream task. With carefully designed prompts and a powerful PLM, the generated dataset is believed to incorporate rich task-specific knowledge.
    
    \item 
    Given pseudo dataset synthesized as above, we then train a \underline{t}iny t\underline{a}sk \underline{m}odel (TAM) to solve the task. TAM can integrate with any task-specific inductive bias and is also order-of-magnitude smaller than PLMs. 

    \item 
    Finally, we perform efficient inference on target task using the trained model. 
    During the whole process, no human annotations are involved, thus the evaluation setting is purely zero-shot.
\end{enumerate}

\paragraph{Pseudo dataset generation}
For a single-sentence classification task $\mathcal{D}$, we aim to generate a pseudo dataset $\mathcal{D}^g = (\mathcal{X}^g, \mathcal{Y}^g) $ with the help of a left-to-right PLM $\mathcal{P}$. 
We first sample a class label $y^g$ from a uniform distribution:
\begin{equation}
y^g \sim \mathbf{U}(y_1, y_2, \ldots, y_k),    
\end{equation}
where $k$ is the number of classes. $y^g$ is then wrapped up into a label-descriptive prompt $\mathcal{T}(y^g)$ to steer the generation of $\mathbf{x}^g$:
\begin{equation}
\label{eq:gen_x}
\mathbf{x}^g \sim \mathcal{P}(\cdot|\mathcal{T}(y^g)).
\end{equation}
Since the parameters of $\mathcal{P}$ is frozen and the generation $\mathbf{x}^g$ for each $y^g$ is deterministic, we can adopt different sampling algorithms (e.g., Top-k sampling \cite{DBLP:conf/acl/LewisDF18} and nucleus sampling \cite{DBLP:conf/iclr/HoltzmanBDFC20}) to increase the diversity of generated dataset. We then pair the generated $\mathbf{x}^g$ with $y^g$ to construct a pseudo training data.  
We show an example about generating a pseudo sentiment classification dataset in Figure \ref{fig:model}(b). The prompt $\mathcal{T}(y^g)$ for a positive label $y^g$ is "\textit{The movie review in positive sentiment is: "}". With the sampling strategies, this prompt steers PLMs to generate multiple 
sentence ending with another quotation mark, e.g., "\textit{A deep and meaningful movie."}" or "\textit{Good film!!!"}". 

For sentence-pair classification tasks, we need to generate two sequences that bear certain relationships (e.g., premise and hypothesis in NLI, context and question in QA). We decompose the generation into two steps: (i) We first generate and/or sample a conditional context $\mathbf{c}^g$ (e.g., $\mathbf{c}^g$ represents premise in NLI and context in QA). The context $\mathbf{c}^g$ is then concatenated with a sampled label $y_g$ and transformed into a prompt $\mathcal{T}(\mathbf{c}^g, y_g)$. (ii) Giving the prompt $\mathcal{T}(\mathbf{c}^g, y_g)$, we can now generate the other sentence $\mathbf{x}^g$ (e.g., hypothesis in NLI and question in QA) as in Equation (\ref{eq:gen_x}). 
In current implementation, we sample $\mathbf{c}^g$ from an unlabeled corpus. But $\mathbf{c}^g$ can also be generated following procedure of generation for single-sentence classification task. Since there could be no predefined label set for extractive QA task, we use publicly available spaCy\footnote{https://spacy.io/} toolkit to annotate entities, and then uniformly select an entity as $y^g$.
Finally, the generated sentence-pair and label can form the pseudo dataset $\mathcal{D}^g = (\mathcal{C}^g, \mathcal{X}^g, \mathcal{Y}^g)$. We elaborate details on prompts chosen for each task in Section \ref{sec:prompt}.

\paragraph{Pseudo-supervised training}
With the pseudo dataset $\mathcal{D}^g$, we train a tiny task model \smallmodel to conduct the given task. This procedure is highly flexible, meaning that we can use any model architecture, loss function, and training strategy. In this work, we primarily focus on the overall framework, thus we leave the tuning of these components for future work. Under the zero-shot learning setting, it should be noted that we have no access to the standard validation set. Therefore, we use a portion (e.g., 10\%) of the pseudo dataset as the validation set for model selection. 

\paragraph{Zero-shot evaluation}
Finally, we conduct inference on the trained \smallmodel model. As \smallmodel is order-of-magnitude smaller than PLM, it is able to perform extremely efficient inference.

\begin{table*}[t]
\centering
\scalebox{0.8}{
\begin{tabular}{ll|ll|cccccc}
\toprule
\textbf{PLM} & \textbf{TAM} & \textbf{\#Param} & \textbf{Setting} & \textbf{IMDb} & \textbf{SST-2} & \textbf{SQuAD} & \textbf{AdversarialQA} & \textbf{QNLI} & \textbf{RTE} \\
\hline
 \#\textit{Gold Data}& & &\multirow{3}{*}{{\textsc{Supervised}}} & 25k & 6.7k & 87k & 30k & 105k & 2.5k\\
 \multirow{2}{*}{-} & DistilBERT & 66M & & 87.24 & 89.68 & 76.28/84.67 & 18.6/29.85 & 88.05 & 58.12 \\
 & LSTM & $\sim$7M &  & 84.60 & 76.30 & 41.86/57.22 & 5.37/11.86 & 69.00 & 54.87 \\
 \hline
\multirow{3}{*}{GPT2} & - & 117M & \textit{\textsc{Prompting}} & 51.52 & 52.52 & 0.80/4.93 & 0.37/2.58 & 50.60 & \textbf{52.70} \\
 & DistilBERT & 66M & \multirow{2}{*}{\textit{\textsc{ZeroGen}}} & \cc\textbf{73.24} & \cc\textbf{80.39} & \cc\textbf{16.44/21.83} & \cc\textbf{5.20/8.26} & \cc\textbf{55.32} & 50.54 \\
 & LSTM & $\sim$7M &  & \cc{69.60} & \cc{70.40} & \cc{4.94/8.53} & \cc{1.00/3.83} & \cc{51.03} & 49.10 \\
 \hline
\multirow{3}{*}{GPT2-Large} & - & 762M & \textit{\textsc{Prompting}} & 80.20 & \textbf{87.84} & 3.53/10.78 & 1.47/5.16 & 55.10 & 54.51 \\
 & DistilBERT & 66M & \multirow{2}{*}{\textit{\textsc{ZeroGen}}} & \cc\textbf{83.56} & 85.44 & \cc\textbf{23.87/29.82} & \cc\textbf{5.93/9.63} & \cc\textbf{69.32} & \cc\textbf{58.48}$^{*}$ \\
 & LSTM & $\sim$7M &  & 78.20 & 75.10 & \cc{8.01/12.77} & \cc{2.33/5.24} & 51.27 & \cc{56.68}$^{*}$ \\
 \hline
\multirow{3}{*}{GPT2-XL} & - & 1.5B & \textit{\textsc{Prompting}} & 80.64 & \textbf{89.22} & 4.61/13.32 & 2.13/6.30 & 60.60 & 57.04 \\
 & DistilBERT & 66M & \multirow{2}{*}{\textit{\textsc{ZeroGen}}} & \cc\textbf{84.28} &  {87.27} & \cc\textbf{25.50/31.53} & \cc\textbf{6.33/9.96} & \cc\textbf{71.19} & \cc\textbf{59.93}$^{*}$ \\
 & LSTM & $\sim$7M &  & 79.80 & 78.40$^{*}$ & \cc{12.35/18.66} & \cc{3.23/6.34} & 52.26 & \cc{58.85}$^{*}$ \\
 \bottomrule
\end{tabular}}
\caption{Evaluation results for \textsc{ZeroGen} at three  different scales of PLM and two different scales of TAM.
The \textsc{ZeroGen} results that outperform \textsc{Prompting} using the same PLM are in \hl{grey}, and the best result for each task using the same PLM is \textbf{bolded}. $^{*}$ indicates that the result of TAM under \textsc{ZeroGen} framework outperforms the same TAM under \textsc{Supervised} framework. 
We report the average number of parameters (i.e., 7M) for LSTM-based models among different tasks.
The scale of the synthetic dataset is 200k for each task. Results on larger PLMs are reported in \S \ref{sec:opt}.}
\label{tab:main}
\end{table*}

\section{Experiments}
\subsection{Setup}
We perform experiments across three different tasks including six different NLP datasets. The detailed experimental setup (i.e., Implementation Details) are described in Appendix \ref{ap:exp}.

\paragraph{Datasets}
We consider two Text Classification datasets (i.e., SST-2 \cite{DBLP:conf/emnlp/SocherPWCMNP13} and IMDb \cite{DBLP:conf/acl/MaasDPHNP11}), two Natural Language Inference datasets (i.e., QNLI \cite{DBLP:conf/emnlp/RajpurkarZLL16} and RTE \cite{DBLP:conf/mlcw/DaganGM05,haim2006second,DBLP:conf/acl/GiampiccoloMDD07,DBLP:conf/tac/BentivogliMDDG09}), and two Question Answering datasets (i.e., SQuAD1.1 \cite{DBLP:conf/emnlp/RajpurkarZLL16} and AdversarialQA \cite{DBLP:journals/tacl/BartoloRWRS20}). The number of training examples for SST-2 and RTE is 6.9k and 2.5k, which can be considered as low resource compared with IMDb (25k), QNLI (105k), SQuAD (87k) and AdversarialQA (30k). We adopt Exact-Match (EM) and $F_1$ as the metrics for question answering tasks and Accuracy for other tasks.  

\paragraph{Baselines}
We compare \textsc{ZeroGen} framework with two baselines: \textbf{(1)} \textsc{Prompting}. The prompt-based zero-shot learning framework via PLMs. We use GPT2 (117M), GPT2-large (762M), and GPT2-XL (1.5B) \cite{radfordlanguage} via the HuggingFace Transformers library \cite{DBLP:journals/corr/abs-1910-03771}. 
\textbf{(2)} \textsc{Supervised}. The TAMs are trained on standard dataset (i.e., human annotations).
Regarding model architecture of TAMs, we use two types of model for each task: a LSTM-based model (i.e., BiLSTM \cite{hochreiter1997long} for TC and NLI tasks, and BiDAF \cite{DBLP:conf/iclr/SeoKFH17} for QA task), and a tiny pre-trained model (i.e., DistilBERT \cite{DBLP:journals/corr/abs-1910-01108}).

\paragraph{Evaluation Strategy}
Due to restricted test set access for some datasets (i.e., SQuAD1.1 and SST-2), we held out a small subset (i.e., 10\%) of the training set for validation for model trainined in \textsc{Supervised} setting, and report results on the validation set. 
For models trained with synthetic dataset in \textsc{ZeroGen} framework, we also use a portion (i.e., 10\%) as the validation set, without accessing to original validation set.
For \textsc{Prompting}, we directly evaluate on the original validation set.

\subsection{\textsc{ZeroGen} vs. \textsc{Prompting}}

\begin{figure*}[t]
\centering
\includegraphics[width=6.3in]{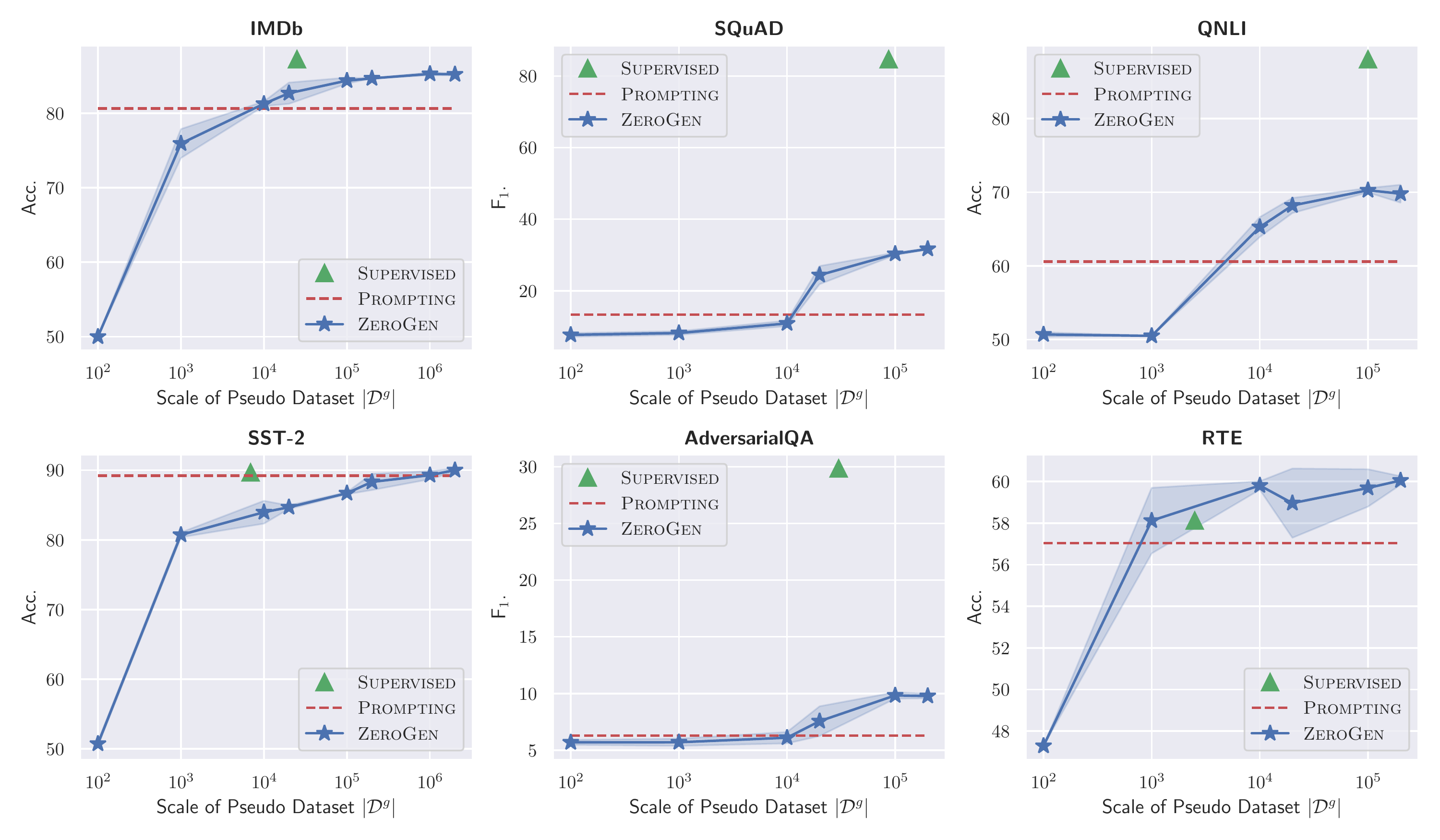}
\caption{Results for comparing various scales of synthetic datasets on different tasks. 
We use GPT2-XL as PLM and DistilBERT as TAM. 
Dots with star marker and error bars are the average performance and the standard deviation over 3 runs,
respectively.}
\label{fig:scale}
\end{figure*}

\label{sec:main}

Table \ref{tab:main} compares \textsc{ZeroGen} with \textsc{Prompting} framework. We observe that \proposed significantly outperforms \textsc{Prompting} on most datasets we evaluated, and this superiority is consistent across different PLM generators and \smallmodels. In particular, when using DistilBERT as \smallmodel, we find that among 18 (3 generators $\times$ 6 tasks) head-to-head comparison with \textsc{Prompting}, \proposed achieves better performance in 15 cases\footnote{Note that with careful prompt design and selection, our \textsc{Prompting} baseline achieves an accuracy of 89.22\% with GPT2-XL on SST-2 dataset, substantially higher than the previous best results (i.e., 87.4\% \cite{DBLP:conf/emnlp/HoltzmanWSCZ21})}. The reasons for the superior performance are mainly two-folds: 1) compared with the general purpose generation model, task-specific classification model may encourage a more deterministic decision boundary, which shares the same spirits with  entropy minimization \cite{grandvalet2006entropy} or self-training \cite{lee2013pseudo}, and 2) classification tasks benefit from the inductive bias in the architecture. For example, method that predicts the start and end positions greatly narrows down the searching space for extractive question answering tasks, in comparison with free-generation on the vocabulary space.

Besides the superior effectiveness in zero-shot learning, it's also worth noting that \proposed is also quite efficient. \proposed can achieve comparable (LSTM) and even better (DistilBERT) performance than \textsc{Prompting}, using more than 200 times and 20 times fewer parameters, respectively. 
Nowadays, with increasingly larger pre-trained language models (e.g., 175B GPT-3 \cite{DBLP:conf/nips/BrownMRSKDNSSAA20}, 1571B Switch-C  \cite{DBLP:journals/corr/abs-2101-03961}), the advantage of \proposed becomes even more pronounced. The gigantic PLMs can improve the quality of synthesized dataset and lead to better zero-shot performance. Meanwhile, the \smallmodel can remain light-weighted for efficient inference and serving.

Furthermore, when scaling up PLMs, we observe continuous performance boost for both \textsc{Prompting} and \textsc{ZeroGen}. This indicates that larger-scale PLMs might have been trained to store more knowledge that is useful for generating accurate dataset for a task.

\subsection{\textsc{ZeroGen} vs. \textsc{Supervised}}
\label{sec:scale}

It's commonly accepted that the zero-shot performance of a NLP model can lag way behind its fully-supervised performance (trained on human annotations). However, we find that \proposed even outperform its \textsc{Supervised} counterpart on SST-2 and RTE datasets (highlighted with $^*$ in the Table~\ref{tab:main}). Our conjecture is the size of the datasets are the key factor. \proposed automatically generates much more data as supervision during training (i.e., 200k synthesized samples vs. 6.9k/2.5k human annotations). These results are encouraging because they suggest that: (i) \proposed is quite effective in low-resource scenario; (ii) it's possible to synthesize training samples to approximate human-annotations in a fully unsupervised manner. 

We further investigate if we can trade data volume in exchange for zero-shot performance in \proposed. Our results are shown in Figure~\ref{fig:scale}.

Overall, we find that the final performance improves continuously as the amount of data grows, despite diminishing returns. We find that generating 10k of training samples leads to better performance than \textsc{Prompting} method on most datasets. In addition, by increasing data size, we find that \proposed even outperforms the \textsc{Supervised} baseline on SST-2 and RTE. But still, on some datasets examined (e.g., SQuAD, QNLI), there remain a performance gap between \proposed and \textsc{Supervised}.

\begin{table}[t]
\centering
\scalebox{0.75}{
\begin{tabular}{ll|ccc}
\toprule
\textbf{TAM} & \textbf{Strategy} & \textbf{IMDb} & \textbf{SQuAD} & \textbf{QNLI} \\
\hline
\multirow{5}{*}{DistilBERT} & Greedy & 74.40 & 24.19/31.16 & 63.59 \\
 & Top k=5 & 79.50 & \textbf{25.68/32.32} & \textbf{71.83} \\
 & Top k=40 & \textbf{83.70} & 24.70/31.21 & 70.40 \\
 & Top k=80 & \textbf{83.70} & 24.14/30.89 & 69.83 \\
 & Nucleus p=0.9 & 83.40 & 23.93/30.62 & 70.55 \\
 \hline
\multirow{5}{*}{LSTM} & Greedy & 54.70 & 10.76/16.24 & 53.14 \\
 & Top k=5 & 57.30 & \textbf{13.22/19.45} & \textbf{54.68} \\
 & Top k=40 & \textbf{72.24} & 10.11/15.56 & 51.24 \\
 & Top k=80 & 71.10 & 9.63/14.70 & 51.38 \\
 & Nucleus p=0.9 & 71.80 & 8.83/13.89 & 51.16 \\
 \bottomrule
\end{tabular}}

\caption{Results for comparing different decoding methods with selected parameters of each method. The best result under different decoding methods is \textbf{bolded}.
}
\label{tab:decode}
\end{table}
\subsection{\proposed as Text Generation Evaluator}
\label{sec:decode}
The quality of the synthesized text is the key to the performance of the downstream tasks. \proposed can thus be seen as an indirect measure of the generation models and algorithms.
It is a commonly accepted belief that the quality of generated text should be in an ascending order in GPT-2, GPT2-Large, and GPT2-XL, due to the growing in the parameter size. We find this trend is well aligned in the downstream application performance (Table~\ref{tab:main}).

Besides the model, another important aspect in text generation is its decoding algorithm, where the goal is to achieve better diversity without the text quality (e.g., fluency, coherence, and correctness). We show that how the trade-off between \emph{diversity} and \emph{correctness} is reflected in the framework of \proposed.

\paragraph{Overall Results}
Sampling strategies (e.g., top-k sampling and nucleus sampling) are known to be able to generate text with a higher degree of diversity than other decoding strategies (e.g., greedy search) \cite{DBLP:conf/acl/LewisDF18,DBLP:conf/iclr/HoltzmanBDFC20}. 
Empirical results in Table \ref{tab:decode} demonstrate that a more diverse decoding strategy 
does not always ensure better performance on downstream tasks.
For example, the results of the nucleus sampling strategy, which is considered to generate the most diverse data, achieves a performance nearly 6\% and 3\% lower than the best decoding strategy on both the SQuAD and QNLI datasets, respectively, while greedy decoding strategy could obtain better results than some sampling strategy (e.g., top-k=40, top-k=80 and nucleus sampling). In contrast, all sampling strategies are superior to the greedy decoding strategy on the IMDb dataset. 
Regarding the inconsistent better downstream performance of more diverse decoding strategies, we hypothesize that diversity may come at a price, such as generating samples not pertain to the class described in the prompt. Therefore, we assess the quality of a dataset from two perspectives: \textit{Diversity} and \textit{Correctness} for quantitative analysis of different datasets.

\begin{table}[t]
\centering
\scalebox{0.85}{
\begin{tabular}{l|ccc}
\toprule
\multicolumn{4}{c}{\textbf{\textsc{Diversity}}} \\
\hline
\textbf{Strategy} & \textbf{IMDb} & \textbf{SQuAD} & \textbf{QNLI} \\
\hline
Oracle & 0.30 & 0.14 & 0.14 \\
\hline
Greedy & 0.92 & 0.55 & 0.54 \\
Top-k=5 & 0.59 & 0.29 & 0.28 \\
Top-k=40 & \textbf{0.25} & 0.17 & \textbf{0.14} \\
Top-k=80 & 0.20 & 0.17 & 0.12 \\
Nucleus p=0.9 & 0.15 & \textbf{0.16} & {0.11} \\
\bottomrule
\end{tabular}}
\scalebox{0.85}{
\begin{tabular}{l|ccc}
\toprule
\multicolumn{4}{c}{\textbf{\textsc{Correctness}}} \\
\hline
\textbf{Strategy} & \textbf{IMDb} & \textbf{SQuAD} & \textbf{QNLI} \\
\hline
Oracle & 92.42 & 95.37 & 92.96 \\
\hline
Greedy & 99.67 & \textbf{31.07} & \textbf{83.18} \\
Top-k=5 & \textbf{94.46} & 18.74 & 74.03 \\
Top-k=40 & {84.91} & 14.57 & 64.31 \\
Top-k=80 & 84.11 & 13.74 & 62.94 \\
Nucleus p=0.9 & 82.53 & {13.30} & {63.37} \\
\bottomrule
\end{tabular}}

\caption{Diversity and Correctness evaluation of generated datasets under different decoding strategies.
Oracle refers to the standard dataset with human annotations. 
Diversity is measured by Self-BLEU4, while Correctness is measured by a well-trained RoBERTa-Large model with standard dataset.
}
\label{tab:quality}
\vspace{-2mm}
\end{table}

\paragraph{Diversity} 
We follow previous work \cite{DBLP:conf/iclr/HoltzmanBDFC20} and compute Self-BLEU \cite{DBLP:conf/sigir/ZhuLZGZWY18} as a metric of diversity. Self-BLEU is calculated by computing the BLEU score of each generated text using all other generations in the evaluation set as references\footnote{Specifically, we randomly sample 1000 generations, each of which is compared with all 999 other generations as references.}.
A lower Self-BLEU score implies higher diversity. 
We report 4-gram based Self-BLEU in the first part of Table \ref{tab:quality}. 
We find that decoding strategies such as top-k and nucleus sampling lead to more diverse generations. 
This finding is consistent with previous works \cite{DBLP:journals/corr/LiMJ16,DBLP:journals/corr/VijayakumarCSSL16,DBLP:conf/iclr/WelleckKRDCW20,DBLP:conf/iclr/HoltzmanBDFC20}.

\paragraph{Correctness} 
Different from the vanilla generation scenario that ends with the generated text, we use the generated text as training dataset for another small model. Therefore, \proposed requires a more emphasis on the correctness of generated text, i.e., whether the generated text pertain to the corresponding class described in the prompt. 
To access the correctness of a synthetic dataset, we first train a RoBERTa-Large \cite{DBLP:journals/corr/abs-1907-11692} model on the standard training dataset, which is then used as a validator to evaluate the synthetic dataset. 
In summary, we find a tradeoff between diversity and correctness, i.e., greater diversity leads to lower correctness. 
We notice even deteriorated outcomes by increasing k, while greedy search achieves the highest performance in terms of correctness. 
These results reflect those of \citet{DBLP:conf/emnlp/MassarelliPPORP20} who also found a tradeoff between factuality and diversity, i.e., while decoding strategies such as top-k and nucleus sampling lead to less repetitive generations, they also produce less verifiable text.
Besides, among different tasks, we find the correctness on oracle datasets are similar (i.e., all larger than 90\%), while that varies substantially on synthetic datasets (i.e., up-to 94.46\% on IMDb and merely 31.07\% on SQuAD). Compared with generating datasets for single text classification tasks (e.g., IMDb), where the PLM only needs to consider a single condition (i.e., label), generating for text-pair tasks requires PLMs to consider multiple conditions synchronously (e.g., answer and context when generating question), which makes it more difficult to control the correctness of the generated sample. This possibly explains the observed variance among tasks.


\paragraph{Human Evaluation}\label{appendix:human_eval}
We report the human evaluation results in Table~\ref{tab:human_evaluation}. The quality of generated data is measured by the \textit{correctness} and \textit{naturalness} metrics. The \textit{correctness} measures whether the label is correct and the content is relevant to the task topic (e.g. movie review for IMDb). The \textit{naturalness} measures whether the generated text
is fluent and similar to human-generated text. We invite 4 experts to participate in the evaluation and each participant is randomly assigned 25 generated samples (100 samples in total) for each decoding strategy.  Table~\ref{tab:human_evaluation} report the mean scores. The results show that greedy search achieves the highest performance
in terms of correctness, which is consistent with the automatic evaluation using Roberta-Large. However, in terms of naturalness/fluency, the greedy search performs the worst. The top-k and nucleus decoding strategies can generate a more fluent context.
\begin{table}[t]
\centering
\scalebox{0.75}{
\begin{tabular}{l|cc|c}
\toprule
\textbf{Method }&\textbf{Correctness} & \textbf{Naturalness} & \textbf{\proposed}\\
\hline 
Oracle  &  0.92	& 4.46 & 87.2\\
\hline
Greedy &  0.91	& 3.55  & 74.4\\
Top k=40 & 0.72	& 3.75 & 83.7\\
Nucleus p=0.9 & 0.81 & 3.89 & 83.4\\
\bottomrule
\end{tabular}}
\caption{Human evaluation on IMDb gold and synthetic dataset using different decoding strategies. We also show the TAM performance to show the ability of \proposed as an evaluator.}
\label{tab:human_evaluation}
\end{table}

\begin{table*}[t]
\centering
\renewcommand\arraystretch{1.2}  
\scalebox{0.7}{
\begin{tabular}{lllllcc}
\toprule
\textbf{Setting} & \textbf{Id} & \textbf{Prompt}& \textbf{Label word} <Y>  & \textbf{Prompt Type} & \textbf{IMDb} & \textbf{SST-2} \\
\hline
\multirow{4}{*}{\textit{\textsc{Prompting}}} & $P_1$ 
& \makecell[l]{"<X>" \\It was <Y>} 
& great/terrible & - &51.32 & 51.83\\ 
\cline{2-7}
& $P_2$ 
& \makecell[l]{
<Y> Movie Review: "<X>"} 
& Positive/Negative & Control code & 60.36 & 52.75\\ 
\cline{2-7}
& $P_3$ 
& \makecell[l]{Task: Write a review for a <Y> movie.
\\
Review:  "<X>" } 
 & good/bad & \makecell[l]{Control code with \\ task description} & 54.20 & 53.50\\ 
\cline{2-7}
& $P_4$ 
& \makecell[l]{The movie review in <Y> sentiment is "<X>"} 
 & \multirow{2}{*}{positive/negative} & \multirow{2}{*}{Natural language style} & 80.64 & 89.22\\ 
& $P_5$ 
& \makecell[l]{The <Y> movie review is "<X>"} & & & 67.60 & 72.36 \\ 
\hline
\multirow{5}{*}{\textit{\textsc{ZeroGen}}} & $P_1^{'}$ &   \makecell[l]{It was a review for a <Y> movie: " }  & great/terrible  & - & \underline{79.36} & \underline{75.00} \\
\cline{2-7}
 & $P_2$ &   \makecell[l]{<Y> Movie Review: "} & Positive/Negative & Control code & \underline{60.88} & \underline{67.43} \\
\cline{2-7}
 & $P_3$ &   \makecell[l]{Task: Write a review for a <Y> movie.
 \\
 Review: " }  & good/bad & \makecell[l]{Control code with \\task specification} & 83.40 & 78.90\\
\cline{2-7}
 & $P_4$ &   \makecell[l]{The movie review in <Y> sentiment is: "} & \multirow{3}{*}{positive/negative} & \multirow{3}{*}{Natural language style} & 81.84 & 86.24 \\
 & $P_4^{'}$ &   \makecell[l]{The movie review in <Y> sentiment for movie "<C>" is: "}  & & & 83.40 & 86.35 \\
& $P_5$ &   \makecell[l]{The <Y> movie review is "} & & & 77.44 & 77.06 \\
\bottomrule
\end{tabular}}
\caption{Results for different prompts on IMDb and SST-2 dev sets. We use GPT2-XL as PLM and DistilBERT as TAM.
<X> and <C> represents the input sentence  and generated movie name respectively. 
$P^{'}$ represents minor revised version of $P$ for text generation. 
For \textsc{ZeroGen}, results are reported using 100k training samples. Scores in \underline{underline} are trained on 10k generated samples, since the prompt ($P_1^{'}$ \& $P_2$) is too weak and cannot generate 100k distinct samples.
}
\label{tab:promptdesign_sst}
\end{table*}
\subsection{Prompt Engineering in \proposed}
\label{sec:prompt}
The design of prompts can have huge impact on \textsc{Prompting}, as pointed by many previous works \citep{mishra2021reframing,wei2022chain}. In this section, we investigate how prompt design instructs text generation and affects \proposed's performance.  We examine three commonly used prompt types: (1) \textit{Control code}~\cite{keskarCTRL2019}, (2) \textit{Control code with task description}, (3) \textit{Natural language style}.  For SST-2 and IMDb, example prompts and corresponding results can be found in Table~\ref{tab:promptdesign_sst} (check Appendix A for other tasks).

From Table~\ref{tab:promptdesign_sst}, we first observe that \textit{natural language} prompts are favored by both \proposed and \textsc{Prompting}, rather than prompts contain control code. We hypothesize the reason being that during the pre-training process, the majority of text data fed to the PLMs are natural language sentences, and therefore the PLMs do not contain enough knowledge in control code.
Moreover, we observe that \proposed is more robust towards different prompts than \textsc{Prompting}: 
for \textsc{Prompting}, a minor change from $P_4$ to $P_5$  will lead to a huge drop  in accuracy (16.2\% drop in IMDb); for \proposed,  applying the same prompt revision, the decrement decreases to 9.4\%.
Compared with \textsc{Prompting} which use prompt to directly instruct label words, \proposed use synthesized data as medium to connect PLM and TAM, thus mitigating the  sharp change brought by prompts.

To further explore the potential of prompting, we investigate the two-stage conditional prompt inspired by \cite{DBLP:conf/emnlp/SchickS21a}. In the running example, based on the task characteristic(to generate a movie review), we first generate movie name using prompt \textit{[Movie: "]} and then prompt sentence using $P_4^{'}$. We can find that with the control of movie name, the generated training corpus is more diverse than using $P_4$. With the desirable correctness (see Table~\ref{tab:quality}), the higher diversity leads to a higher accuracy (from 81.84 to 83.40 in IMDb).

The most suitable prompting type in Question Answering and Natural Language Inference tasks has some differences with Text Classification due to different task characteristics.
For details, please refer to the Appendix \ref{ap:prompt}.

\subsection{\proposed via Larger PLM Generator }
\label{sec:opt}

We further investigates the performance of \textsc{ZeroGen} on a larger PLM (i.e., OPT~\citep{zhang2022opt} with 175B parameters). We find both \textsc{Prompting} and \textsc{ZeroGen} benefit from the larger PLM on hard tasks (i.e., SQuAD). But on relatively simpler text classification tasks, the results degrades. This demonstrates that prompt selection is still important for larger models, and the prompt that suits for one model (e.g., GPT2-XL) may not suit for another (e.g., OPT).
\begin{table}[t]
\centering
\scalebox{0.68}{
\begin{tabular}{ll|ccc}
\toprule
\multicolumn{1}{l}{\textbf{PLMs}} & \textbf{Setting}& \textbf{IMDB} & \textbf{SQuAD} & \textbf{QNLI} \\
\hline
\multirow{3}{*}{GPT2-XL} & \textsc{Prompting}  & 80.64 & 4.61/13.32 & 60.60 \\
 & \proposed-LSTM & 79.80 & 12.35/18.66 & 51.53 \\
 &\proposed-DistilBERT & \textbf{84.28} & \textbf{25.50/31.53} & \textbf{71.19} \\
 \hline
\multirow{3}{*}{OPT} & \textsc{Prompting} & 63.18 & 23.35/39.32 & \textbf{54.51} \\
 & \proposed-LSTM & 73.08 & 21.46/30.06 & 50.76 \\
 & \proposed-DistilBERT & \textbf{79.99} & \textbf{33.27/44.91} & 52.97 \\
 \bottomrule
\end{tabular}}
\caption{Comparison of GPT2-XL (1.5B) and OPT (175B) under the same prompt and decoding strategy.}
\label{tab:opt}
\end{table}

\section{Related Work}
\subsection{Prompt-based Zero-shot Learning}

With manual crafted natural language prompt, large-scale PLMs have shown impressive zero-shot abilities in a wide array of NLP tasks\cite{radfordlanguage,DBLP:conf/nips/BrownMRSKDNSSAA20}. 
However, current prompt-based zero-shot learning can be unstable: the choice of prompt contributes a lot to the final performance. 
This motivates researchers to investigate better ways to automatically search and/or manually construct a proper prompt \cite{DBLP:journals/corr/abs-2012-00955,DBLP:conf/emnlp/ShinRLWS20,DBLP:conf/chi/ReynoldsM21,DBLP:journals/corr/abs-2109-07830}. 
To improve the zero-shot generalization across different prompts, another line of work uses a multitask training mixture made up of a large set of different tasks specified in natural language prompts. This induces a model to better generalize to unseen tasks, as well as being more robust to the wording choices of the prompts. \cite{DBLP:conf/emnlp/KhashabiMKSTCH20,DBLP:conf/emnlp/ZhongLZK21,DBLP:journals/corr/abs-2104-08773,DBLP:journals/corr/abs-2109-01652,DBLP:journals/corr/abs-2110-08207,DBLP:journals/corr/abs-2201-06910}. 
In comparison, we advocate and analyse a new paradigm for prompt-based zero-shot learning via dataset generation, which is complementary to current prompt searching and multi-task pre-training methods.

\subsection{Dataset Generation with PLMs}
Our work also relates to research in generating data with PLMs, which aims to generate a pseudo dataset to enhance model performance.  Early efforts achieve this goal with fine-tuned generative models \cite{DBLP:conf/aaai/Anaby-TavorCGKK20,DBLP:conf/emnlp/PuriSSPC20,DBLP:journals/corr/abs-2003-02245,DBLP:journals/corr/abs-2102-01335}. They first fine-tune the generative models using human annotations, the generated data samples are then combined with human annotations to train the models in a semi-supervised fashion. Supervised data generation methods are also studied for building auxiliary tasks \cite{DBLP:conf/emnlp/VuLLSI21} and dataset creation based on human and machine collaboration \cite{DBLP:journals/corr/abs-2201-05955}.
To reduce the human efforts on data annotation, another line of works explore data generation methods without the need for human annotations. 
\citet{DBLP:journals/corr/abs-2106-06168} uses unsupervised-trained unconditional generative models to synthesize unlabeled data for semi-supervised learning. Without any model training,  
\citet{DBLP:journals/corr/abs-2109-09193} propose to directly use unlabeled in-domain examples as prompts to synthesize high-quality training data.  \citet{DBLP:conf/emnlp/SchickS21a} explore dataset generation method from scratch for semantic textual similarity task. One concurrent work \cite{meng2022generating} studies dataset generation for text classification and natural language inference tasks. 
In comparison, we take the dataset generation framework to the extreme, i.e., consider extremely tiny edge models (e.g., LSTM), explore boarder NLP tasks including question answering, and conduct extensive analysis such as decoding strategies and quality evaluation. 

\section{Conclusion and Future Directions}


In this paper, we study an extreme instance of dataset generation via PLMs for zero-shot learning. 
Without any human annotations, we show that an small LSTM can surpass the zero-shot performance of its PLM counterparts (e.g., GPT2-XL), and even outperform the same model trained with human annotations. Despite the demonstrated effectiveness, 
we discuss several issues we observed when developing \proposed and reveal a substantial room of improvement in future research.

Despite positive results on TC tasks, we find the stability regarding prompt choice of \proposed is still far from satisfactory on NLI tasks. Future work could include multi-task prompt-based pre-training methods \cite{DBLP:journals/corr/abs-2110-08207, DBLP:journals/corr/abs-2109-01652}. 

Furthermore, we observe noisy examples in synthetic dataset on difficult tasks such as NLI and QA, this situation progressively deteriorates when incorporating more diverse decoding strategy (e.g., Nucleus Sampling). Better decoding strategies are needed to ensure the label correctness while preserving the dataset diversity \cite{DBLP:conf/emnlp/MassarelliPPORP20}. Besides, methods that learn from noisy labels can be integrated into the training of the tiny task model \cite{DBLP:journals/corr/abs-2007-08199}. 

We hope this paper can provide contributions for further exploiting dataset-generation-based zero-shot learning with large pre-trained language models.

\section*{Limitations}
Although \textsc{ZeroGen} achieves promising performance under zero-shot learning setting, this choice does come with certain limitations. We find the stability regarding the prompt choice of \textsc{ZeroGen} is still far from satisfactory. \textsc{ZeroGen} underperforms \textsc{Prompting} in some certain selected prompts, and prompt engineering is tough as it's shown a different preference on prompts across various tasks.
Future work may include multi-task prompt-based pre-training methods \citep{DBLP:journals/corr/abs-2110-08207, DBLP:journals/corr/abs-2109-01652} to improve prompt robustness. 

We also observe noisy examples in the synthetic dataset on difficult tasks such as NLI and QA, this situation progressively deteriorates when incorporating a more diverse decoding strategy (e.g., Nucleus Sampling). Better decoding strategies are needed to ensure the label's correctness while preserving the dataset diversity \cite{DBLP:conf/emnlp/MassarelliPPORP20}. Reciprocally, methods that learn from noisy labels can be integrated into the training of the tiny task model \cite{DBLP:journals/corr/abs-2007-08199}.

\section*{Acknowledgement}
We thank the anonymous reviewers whose suggestions helped clarify this work. This work is partially supported by the Shanghai Committee of Science and Technology (Grant No.~21DZ1100100), and the joint research scheme of the National Natural Science Foundation of China (NSFC) and the Research Grants Council (RGC) under grant number N\_HKU714/21.

\bibliography{main}
\bibliographystyle{acl_natbib}

\appendix
\section{Experimental Setup}
\label{ap:exp}

\paragraph{Implementation Details}
For dataset generation, we use Nucleus Sampling \cite{DBLP:conf/iclr/HoltzmanBDFC20} with $p=0.9$ by default as it is considered to be able to generate both fluent and diverse texts\cite{DBLP:conf/iclr/HoltzmanBDFC20}. 
The scale of synthetic dataset is 200k in the main results, and 100k in other analysis experiments. 
Regarding prompt selection, we manually design a series of prompts for each task, and report results on the best prompt for \textsc{Prompting} and \textsc{ZeroGen} framework. 
For NLI tasks, we adopt self-debiasing mechanism with a decay constant of 200 \cite{DBLP:journals/corr/abs-2103-00453} to ensure that each generated text pair is not only a good fit for a given label, but also not a good fit for other labels \cite{DBLP:conf/emnlp/SchickS21a}. We removing overly short/long sentences, and sentences without an ending quotation mark. 

We implement a LSTM-based model and a DistilBERT model as TAM. For LSTM-based model, we use Adam optimizer \cite{DBLP:journals/corr/KingmaB14}, a learning rate of 1e-4, an embedding dim of 100, and a hidden size of 300.
For single sentence classification(TC), we  use 1-layer BiLTSM to encode the sentence and use a linear classifier. For sentence-pair classification(NLI), we use 2-layer BiLTSM to encode the sentences to $v_1, v_2$ and pass the concatenated  [$v_1;v_2;|v_1-v_2|;v_1*v_2$] to a classifier. For QA tasks, we use the 1-layer BiDAF model.
To ensure that TAMs are truly trained from scratch using the synthetic corpus, we random initialize TAMs' embedding without 
using any pre-trained word embeddings (e.g., GloVe \cite{DBLP:conf/emnlp/PenningtonSM14}.
For DistilBERT, we fine-tune on each dataset with Adam optimizer, with a learning rate of 2e-5, a weight decay of 0.01, and other default hyper-parameters as suggested by HuggingFace Transformers library \cite{DBLP:journals/corr/abs-1910-03771}.
We run experiments on a single NVIDIA A100 80G GPU, and generating 200k examples cost 12h on average.

\section{Additional Results on Prompt Design}
\label{ap:prompt}

For Question Answering tasks, the \textit{natural language style} prompt is also the most suitable for both \textsc{Prompting} and \textsc{ZeroGen} settings, achieving the highest scores. However, for Natural Language Inference tasks, the most suitable prompts for QNLI and RTE are different. For RTE, the \textit{natural language style} prompt is best, while the \textit{control code} prompts perform significantly better than \textit{natural language style} prompts in QNLI.

\section{Additional Related Work on Efficient Inference of PLMs}
There is a line of works dedicated to improving the inference efficiency of PLMs, including pruning \cite{DBLP:conf/emnlp/WangWL20,DBLP:conf/rep4nlp/GordonDA20}, low-rank factorization \cite{DBLP:conf/nips/MaZZDHZ019,DBLP:conf/ijcnlp/NoachG20,DBLP:conf/iclr/LanCGGSS20}, quantization \cite{DBLP:conf/nips/ZafrirBIW19,DBLP:conf/aaai/ShenDYMYGMK20,kim2021bert}, knowledge distillation \cite{DBLP:conf/emnlp/JiaoYSJCL0L20,DBLP:journals/corr/abs-1910-01108,DBLP:conf/acl/SunYSLYZ20} and parallel decoding \cite{gu2018non,DBLP:conf/emnlp/GhazvininejadLL19, DBLP:conf/acl/YeGL0Z20}. We refer the readers to \citet{DBLP:journals/corr/abs-2111-05193} for a detailed survey. Concerning privacy, copyright or confidentiality, data-free knowledge distillation (DFKD) \cite{DBLP:journals/corr/abs-2112-15278} has attracted appealing attention in computer vision field as it deals with distilling valuable knowledge from well-trained models without requiring to access to the training data. However, similar approaches for NLP are difficult to work due to discrete character of words. \citet{DBLP:conf/emnlp/RashidLGR21} relax the data-free condition and use out-of-distribution labeled data to train a generator. By contrast, our method generates data with the PLMs (i.e., the teacher), without requiring any pre-defined labeled data. In the literature of knowledge distillation, \proposed framework could produce a student model that achieves superior zero-shot performance the teacher model.

\section{\proposed as Knowledge Distillation}
\proposed can be seen as a dataset-based knowledge distillation framework. We compare vanilla knowledge distillation baselines with \proposed in Table~\ref{tab:kd}. The soft and hard labels are generated by GPT2-XL on the unlabeled training set. The generated labels are used to train a tiny task model for comparison. The superior results on \proposed show that the paradigm can better utilize large PLM by distilling more knowledge into a large amount of input-output pairs, while vanilla knowledge distillation purely distills knowledge into outputs.
\begin{table}[htbp]
\centering
\scalebox{0.7}{
\begin{tabular}{l|cccc}
\toprule
\textbf{TAMs} & \textbf{Supervised}  & \textbf{\textsc{KD-Hard}} & \textbf{\textsc{KD-Soft}} & \textbf{\textsc{ZeroGen}} \\
\hline
LSTM & 84.60 & 75.23 & 68.31 & 79.80  \\
DistilBERT& 87.24 & 82.32 & 80.21  & 84.28 \\
\bottomrule
\end{tabular}}
\caption{Comparison with knowledge distillation (KD) baselines on IMDb. \textsc{KD-Hard} and \textsc{KD-Soft} represent KD baselines using hard labels and soft labels, respectively.  }
\label{tab:kd}
\end{table}

\section{\proposed for Data Augmentation}
We report the results using the synthetic data as augmentation data in Tabel~\ref{tab:aug}. The results show that the zero-shot synthetic data is a good supplement to human-annotated data (gold data) and can improve the model performance. 
\begin{table}[htbp]
\centering
\scalebox{0.8}{
\begin{tabular}{ll|cc}
\toprule
 \textbf{Data} & \textbf{Sample Size}  & \textbf{LSTM} & \textbf{DistilBERT} \\
\hline
Gold & 25,000 & 84.60 & 87.24\\
Gold + \textsc{Aug}-200k & 225,000 & 88.91 & 93.42\\
Gold + \textsc{Aug}-500k & 525,000 & 90.42 & 93.59\\
\bottomrule
\end{tabular}}
\caption{Results of data augmentation on IMDb using  synthetic data. \textsc{Aug}-200k 
 and \textsc{Aug}-500k  represent using 200k and 500k synthetic data respectively. }
\label{tab:aug}
\end{table}

\section{\proposed for Self-improving}
We have shown that a tiny task model can outperform a large PLM after training on the synthetic dataset. A natural question is \textit{"Can PLM improve its own performance after tuning on the dataset generated by itself?"}. We experiment using PLM as TAM and report the results in Table~\ref{tab:tune_plm}. To summarize, we find 1) A
larger TAM can further boosts the performance; 2) PLMs can improve itself by fine-tuning on the dataset generated by its own.
\begin{table}[h]
\centering
\scalebox{0.8}{
\begin{tabular}{ll|ccc}
\toprule
\textbf{PLMs} & \textbf{TAMs} & \textbf{IMDb} & \textbf{SQuAD} & \textbf{QNLI} \\
\hline
\multirow{3}{*}{-} & LSTM & 84.60 & 41.86/57.22 & 69.00 \\
 & DistilBERT & 87.24 & 76.28/84.67 & 88.05 \\
 & GPT2-XL & {95.68} & {76.92/85.48} & {92.88} \\
 \hline
\multirow{4}{*}{{GPT2-XL}} & - & \cc 80.64 & \cc 4.61/13.32 & \cc 60.60 \\
 & LSTM & 79.80 & 12.35/18.66 & 51.53 \\
 & DistilBERT & 84.28 & 25.50/31.53 & 71.19 \\
 & GPT2-XL & \cc \textbf{90.71} & \cc \textbf{25.78/32.13} & \cc\textbf{73.69} \\
 
 \bottomrule
\end{tabular}}
\caption{Results of PLM (i.e., GPT2-XL) fine-tuned with gold (upper) and synthetic (lower) dataset. A larger TAM further boosts the performance, and PLM can improve its own performance after fine-tuning on synthetic dataset by itself (grey blocks).}
\label{tab:tune_plm}
\end{table}

\section{Generated Examples}
\label{ap:examples}
We present some qualitative examples for different tasks in Appendix Table~\ref{tab:examples}. 
Text classification task~(SST-2) is relatively simple and concise, the generated samples generally fit the prompts and sentiment polarity well by using descriptive tokens about the given movie name and positive/negative sentiment.
Take the first case in SST-2 as an example, the generated tokens ``\textit{action-adventure}'' and ``\textit{attractive}'' are the natural continuations for movie name ``\textit{The Spiderwick Chronicles (Movie)}'' and ``\textit{positive}'' sentiment in prompt.
Although natural language inference tasks are complex,  the generated questions~(QNLI) and inferences~(RTE) could respond to different types of prompts and relate to the given contexts~(e.g., the generated question drifts topic for prompt ``\textit{Information:\ldots Question (answer \textbf{not in} above information)}'' in QNLI). 
While the context of the question answering task~(SQuAD) is long and contains a lot of information, \proposed could successfully generate question ``\textit{Who is the one and only true God ?}'' which is used to response to the pre-set answer ``\textit{Jehovah}''.
Overall, these generation examples show that \proposed can generate useful and arbitrary number of training samples that could be used to train TAMs.

\begin{table*}[t]
\centering
\scalebox{0.7}{
\begin{tabular}{llllcc}
\toprule
\textbf{Setting} & \textbf{Id} & \textbf{Prompt} &\textbf{Prompt Type}& \textbf{SQuAD} & \textbf{AdversarialQA} \\
\hline
\multirow{5}{*}{\textit{\textsc{Prompting}}} & $P_1$ 
& \makecell[l]{ Context: "<C>"\textbackslash nQuestion: "<X>"\textbackslash nAnswer: "}& Control code & 2.69/10.90 & 1.33/5.61 \\ 
\cline{2-6}
& $P_2$ 
& \makecell[l]{Task: Generate an answer given the context and question.\\
\textbackslash nContext: "<C>"\textbackslash nQuestion: "<X>"\textbackslash nAnswer: "} & \makecell[l]{Control code with\\ task description}
& 3.81/12.1	& 1.60/5.89\\ 
\cline{2-6}
& $P_3$ 
& \makecell[l]{"<C>"\textbackslash n
The answer to the question "<X>" is: "} & \multirow{4}{*}{Natural language style}
& 4.41/12.60 &	2.00/5.78 \\ 
& $P_4$ 
& \makecell[l]{"<C>"\textbackslash nBased on the above description, the\\ answer to the question "<X>" is: "} 
&  & 6.33/15.49 &	2.27/6.71 \\ 
& $P_5$ 
& \makecell[l]{The context is: "<C>"\textbackslash nThe answer to the question "<X>" is: "} 
& & 4.61/13.32 &	2.13/6.30\\ 
\hline
 \multirow{5}{*}{\textit{\textsc{ZeroGen}}}
& $P_1^{'}$ 
& \makecell[l]{ Context: "<C>"\textbackslash nAnswer: "<Y>"\textbackslash nQuestion: "} 
& Control code & 9.96/9.62 & 2.87/5.33 \\ 
\cline{2-6}
& $P_2^{'}$ 
& \makecell[l]{Task: Generate a question given the context and answer.\\
\textbackslash nContext: "<C>"\textbackslash nAnswer: "<Y>"\textbackslash nQuestion: "}  & \makecell[l]{Control code with\\ task description}
& 5.84/8.70 & 2.77/5.21 \\ 
\cline{2-6}
& $P_3^{'}$ 
& \makecell[l]{"<C>"\textbackslash n"<Y>" is the answer to the question: "} & \multirow{5}{*}{Natural language style}
& 24.55/29.36 &	5.30/8.82 \\ 
& $P_4^{'}$ 
& \makecell[l]{"<C>"\textbackslash nBased on the above description, "<Y>" is the\\ answer to the question: "} 
& & 23.58/29.84 & 5.87/9.58  \\ 
& $P_5^{'}$ 
& \makecell[l]{The context is: "<C>"\textbackslash n"<Y>" is the answer to the following\\ question: "} 
&  &23.93/30.62 & 5.97/10.02  \\ 
\bottomrule
\vspace{1mm}
\end{tabular}}

\scalebox{0.7}{
\begin{tabular}{lllllc}
\toprule
\textbf{Setting} & \textbf{Id} & \textbf{Prompt} & \textbf{Label words} <Y> & \textbf{Prompt Type} &\textbf{QNLI}\\
\hline
\multirow{5}{*}{\textit{\textsc{Prompting}}} & $P_1$ 
& \makecell[l]{Context: "<X>"\textbackslash nQuestion (answer <Y> the context): "} 
& in/not in  & \multirow{2}{*}{Control code
}& 50.51\\ 
& $P_2$ 
& \makecell[l]{Information: "<X>"\textbackslash nQuestion (answer <Y> above information): "} 
&   in/not in & & 50.52\\ 
\cline{2-6}
& $P_3$ 
& \makecell[l]{"<X>"\textbackslash n Based on the above description, the following question is\\ \textsc{[Y1]} and \textsc{[Y2]} be answered: "} 
&  \makecell[l]{clear/not clear\\ can/can not} & \multirow{6}{*}{Natural language style} &60.42  \\ 
& $P_4$ 
& \makecell[l]{The context sentence is: "<X>"\textbackslash nThe question is: "<X>"\textbackslash n\\
The context sentence <Y> the answer to the question. \\} 
& \makecell[l]{contains/\\doesn't contain} & &60.61 \\ 
& $P_5$ 
& \makecell[l]{The context sentence is: "<X>"\textbackslash nThe context sentence <Y> the\\ answer to the following question.\textbackslash nThe question is: "<X>"
} 
& \makecell[l]{contains/\\doesn't contain} & &58.02 \\ 
\hline
 \multirow{6}{*}{\textit{\textsc{ZeroGen}}}
& $P_1$ 
& \makecell[l]{Context: "<X>"\textbackslash nQuestion (answer <Y> the context): "} 
& in/not in &\multirow{2}{*}{Control code
} & 69.82\\ 
& $P_2$ 
& \makecell[l]{Information: "<X>"\textbackslash nQuestion (answer <Y> above information): "} 
&   in/not in & & 70.55\\ 
\cline{2-6}
& $P_3$ 
& \makecell[l]{"<X>"\textbackslash n Based on the above description, the following question is\\ \textsc{[Y1]} and \textsc{[Y2]} be answered: "} 
&  \makecell[l]{clear/not clear\\ can/can not} & \multirow{4}{*}{Natural language style
} &53.71 \\ 
& $P_4^{'}$ 
& \makecell[l]{The context sentence is: "<X>"\textbackslash nThe context sentence <Y> the\\ answer to the following question.\textbackslash nThe question is: "<X>"} 
& \makecell[l]{contains/\\doesn't contain}  & & 55.65 \\ 
\bottomrule
\vspace{1mm}
\end{tabular}}
\scalebox{0.7}{
\begin{tabular}{lllllc}
\toprule
\textbf{Setting} & \textbf{Id} & \textbf{Prompt} & \textbf{Label words} <Y> & \textbf{Prompt Type}& \textbf{RTE}\\
\hline
\multirow{5}{*}{\textit{\textsc{Prompting}}} & $P_1$ 
& \makecell[l]{Task: Write two sentences have the <Y> meaning.\\
\textbackslash nThe first sentence: "<X>"\textbackslash nThe second sentence: "} 
& same/complete different & \makecell[l]{Control code with \\task description} & 52.71\\ 
\cline{2-6}
& $P_2$ 
& \makecell[l]{Based on the fact that "<X>", it is <Y> that: "} 
&   correct/not correct & \multirow{4}{*}{Natural language style}  & 51.26 \\ 
& $P_3$ 
& \makecell[l]{Suppose "<X>", we <Y>  infer that: "} 
&  \makecell[l]{can/ can not} & & 52.71  \\ 
& $P_4$ 
& \makecell[l]{The context is: "<X>"\textbackslash nBased on the context, it is <Y>  that: "} 
& correct/not correct & &54.51  \\ 
& $P_5$ 
& \makecell[l]{"<X>"\textbackslash nBased on the above description, the following sentence\\ is definitely <Y> : "} 
& correct/wrong & &57.04 \\ 
\midrule
\multirow{5}{*}{\textit{\textsc{ZeroGen}}} & $P_1$ 
& \makecell[l]{Task: Write two sentences have the <Y> meaning.\\
\textbackslash nThe first sentence: "<X>"\textbackslash nThe second sentence: "} 
& same/complete different & \makecell[l]{Control code with \\task description} & 47.29\\ 
\cline{2-6}
& $P_2$ 
& \makecell[l]{Based on the fact that "<X>", it is <Y> that: "} 
&   correct/not correct & \multirow{4}{*}{Natural language style}  & 57.76 \\ 
& $P_3$ 
& \makecell[l]{Suppose "<X>", we <Y>  infer that: "} 
&  \makecell[l]{can/ can not} &  &49.46 \\ 
& $P_4$ 
& \makecell[l]{The context is: "<X>"\textbackslash nBased on the context, it is <Y>  that: "} 
& correct/not correct &  &61.73 \\ 
& $P_5$ 
& \makecell[l]{"<X>"\textbackslash nBased on the above description, the following sentence\\ is definitely <Y> : "} 
& correct/wrong & & 62.82 \\ 
\bottomrule
\end{tabular}}
\caption{Results for different prompts on Question Answering and  Natural language Inference tasks. Results are reported on dev set. For Question Answering, "<C>" represents the input context, "<X>" represents the input question, "<Y>" represents the generated answer. "\textbackslash n" represents the newline symbol. $P^{'}$ represents minor revised version of $P$ for text generation.}
\label{tab:promptdesign_others}
\end{table*}

\begin{table*}[t]
\centering
\scalebox{0.8}{
\begin{tabular}{p{1.1cm}|p{13.5cm}p{4cm}}
\toprule
\textbf{Task} & \textbf{Prompt} & \textbf{Sampled Generation}\\
\hline
\multirow{2}{*}{SST-2}
& The movie review in \textbf{positive} sentiment for movie "\textbf{The Spiderwick Chronicles (Movie)}" is: " & It's an action-adventure movie about a crime fighter who must face an even bigger foe, then the end of the world, all in a funny, attractive, and moving way. \\
\cline{2-3}
& The movie review in \textbf{negative} sentiment for movie "\textbf{Fist Fight}" is: " & The action scenes are laughable, the characters lack personality, and the plot is boring. \\
\hline
\multirow{2}{*}{QNLI}
& Information: "\textbf{In September 1695, Captain Henry Every, an English pirate on board the Fancy, reached the Straits of Bab-el-Mandeb, where he teamed up with five other pirate captains to make an attack on the Indian fleet making the annual voyage to Mocha.}" \textbackslash n Question (answer \textbf{in} above information): " & Why did each of these captains enter the Straits of Bab-el-Mandeb? \\
\cline{2-3}
& Information: "\textbf{Gaddafi remained the government's public face, with the identities of the other RCC members only being publicly revealed on 10 January 1970.}" \textbackslash n Question (answer \textbf{not in} above information): " & Was the financing received from the international revolution from Germany (4)? \\
\hline
\multirow{2}{*}{RTE}
& "\textbf{Today about 75\% of people live in cities or towns. Art and culture are an important part of France.}" \textbackslash n Based on the above description, the following sentence is definitely \textbf{correct}: "& Today about 75\% of people live in cities and towns. \\
\cline{2-3}
& "\textbf{There are four identified subtypes of Ebola virus. Three of the four have caused disease in humans: Ebola-Zaire, Ebola-Sudan, and Ebola-Ivory Coast. The fourth, Ebola-Reston, has caused disease in non-human primates, but not in humans.}" \textbackslash n Based on the above description, the following sentence is definitely \textbf{wrong}: "& The most recently documented case was a Sierra Leonean man who died in 2015. \\
\hline
\multirow{1}{*}{SQuAD}
& The context is: "\textbf{On the other hand, in his study on nine of \"the Bibles most widely in use in the English-speaking world\", Bible scholar Jason BeDuhn, Professor of Religious Studies at the Northern Arizona University, wrote: \u201cThe NW [New World Translation] emerges as the most accurate of the translations compared.\u201d Although the general public and many Bible scholars assume that the differences in the New World Translation are the result of religious bias on the part of its translators, BeDuhn stated: \u201cMost of the differences are due to the greater accuracy of the NW as a literal, conservative translation of the original expressions of the New Testament writers.\u201d He added however that the insertion of the name Jehovah in the New Testament \"violate[s] accuracy in favor of denominationally preferred expressions for God.}" \textbackslash n "\textbf{Jehovah}" is the answer to the following question: "& Who is the one and only true God ?
\\ 
\bottomrule
\end{tabular}}

\caption{Generated examples for each task. We omit the example for IMDb and Adversarial QA tasks since we use the exactly same prompt as SST-2 and SQuAD task, respectively. The input conditions in each prompt are \textbf{bolded}.}
\label{tab:examples}
\end{table*}
\end{document}